\documentclass[10pt,journal,compsoc]{IEEEtran}
\usepackage{epsfig}
\usepackage{graphicx}
\usepackage{graphics} % for pdf, bitmapped graphics files
\usepackage{amsmath}
\usepackage{amssymb}
\usepackage{color, soul}
\usepackage{subcaption}

\begin{document}
\title{Fully Using Classifiers for Weakly Supervised Semantic Segmentation\\with Modified Cues}

\author{Ting~Sun,
	    Lei Tai,
        Zhihan~Gao,
        Ming~Liu,
        and~Dit-Yan~Yeung% <-this % stops a space
\IEEEcompsocitemizethanks{\IEEEcompsocthanksitem Ting~Sun, Lei Tai and Ming~Liu are with the Department
of Electric and Computer Engineering, Hong Kong University of Science and Technology, Clear Water Bay, Hong Kong.\protect\\
E-mail: tsun@ust.hk, eelium@ust.hk
\IEEEcompsocthanksitem Zhihan~Gao and Dit-Yan~Yeung are with the Department
of Electric and Computer Engineering, Hong Kong University of Science and Technology, Clear Water Bay, Hong Kong.\protect\\
E-mail: zhihan.gao@connect.ust.hk, dyyeung@cse.ust.hk}% <-this % stops an unwanted space
}

\markboth{Journal of \LaTeX\ Class Files,~Vol.~14, No.~8, August~2015}%
{Shell \MakeLowercase{\textit{et al.}}: Bare Demo of IEEEtran.cls for Computer Society Journals}


\IEEEtitleabstractindextext{%
\begin{abstract}
This paper proposes a novel weakly-supervised semantic segmentation method using image-level label only.  The class-specific activation maps from the well-trained classifiers are used as cues to train a segmentation network.
The well-known defects of these cues are coarseness and incompleteness.
We use super-pixel to refine them, and fuse the cues extracted from both a color image trained classifier and a gray image trained classifier to compensate for their incompleteness.
The conditional random field is adapted to regulate the training process and to refine the outputs further.
Besides initializing the segmentation network, the previously trained classifier is also used in the testing phase to suppress the non-existing classes.  Experimental results on the PASCAL VOC 2012 dataset illustrate the effectiveness of our method.
\end{abstract}

\begin{IEEEkeywords}
weakly-superviesed, semantic segmentation, super-pixel.
\end{IEEEkeywords}}

\maketitle

\IEEEdisplaynontitleabstractindextext

% \fi
\IEEEpeerreviewmaketitle

\section{Introduction}
\label{sec:introduction}
%\section{Introduction}
%\label{sec:intro}
Semantic segmentation is one of the most fine-grained and challenging computer vision tasks.  The goal is to label every pixel in the image with one of the several predetermined categories.  Given enough manually labelled training data, current deep learning methods achieve impressive performance \cite{chen2018deeplab}.  However, the pixel-level annotations are laborious and expensive to collect, so weakly-supervised semantic segmentation is receiving growing attention and will have a significant impact on this area.

The existing weakly supervised semantic segmentation methods are mainly based on various annotations including bounding boxes, user scribbles, web images, saliency masks and image-level labels.  To clarify the problem setting and for a fair comparison, we consider the annotation needed for the overall segmentation system including the supervision of the intermediate modules. A segmentation network may require both image-level class labels and super-pixels as supervision. If the super-pixel is generated based on prior knowledge of the image-level properties, we consider this method as an image-level weak supervision method. However, if the super-pixel generator is trained with annotations other than image labels, these will be counted as additional supervision required by the system.
Similarly, saliency, i.e., the foreground/background mask, is very helpful for segmentation, and we found that in the PASCAL VOC 2012 augmented dataset \cite{everingham2010pascal, mark2015ijcv, hariharan2011semantic}, there are 12030 images in total, and 7673 of them contain only one class, i.e., for over 63\% of the cases the saliency is equivalent to the segmentation mask.  For a fair comparison, the annotation needed to train the saliency detector should be counted in the supervision of the overall segmentation system.

Since image level annotation, i.e., class labels, are abundant and relatively cheap to collect, we build a segmentation system that only requires image-level class labels during training.  Our work is mainly inspired by SEC~\cite{kolesnikov2016seed},  where the class-specific activation maps are extracted from a well-trained classifier through class activation maps (CAM)~\cite{zhou2016learning}. The active regions are served as high confidence cues to train a segmentation network.  The main contribution of SEC~\cite{kolesnikov2016seed} is to define the loss function that contains three terms, i.e., the seeding loss to impose the activation in the cue region, the constrain-to-boundary loss to encourage the consistency of the segmentation masks before and after applying conditional random field (CRF)~\cite{lafferty2001conditional,krahenbuhl2011efficient}, and the expansion loss which penalizes the classification error.  Among the three loss terms, the seeding loss plays a crucial role, and our idea is to improve these segmentation cues.  We also modify the training and testing procedure.

It is known that the activation in the hidden layer feature maps of a well-trained classifier localizes the region of interest, but it has the following limitations for the segmentation tasks: 1) it is too coarse, and 2) it only highlights the discriminative region, which is not necessary to be the whole object.  We refine the activation cues from a well-trained classifier by snapping it to the super-pixels in the image.  We also train another classifier using gray images to extract cues that capture more structural information, then fuse the cues from the two classifiers to train the segmentation network.  We omit the classification loss in the original SEC since we found that it makes the mask over discriminative and decreases the performance if the weight of this loss term is not carefully chosen.  After generating cues, we preserve the classifier, and not only use its weights to initialize the segmentation network, but also use its prediction in the testing phase to amend the segmentation mask by suppressing the nonexistent classes.

%; 3) the activation is class-agnostic, which makes it non-trivial for multi-class segmentation, but this problem is neatly solved by \cite{zhou2016learning,selvaraju2016grad}.

To summarise, we contribute a weakly-supervised semantic segmentation system that only requires image labels for training.  The novel modules of the system are listed as follows:

\begin{enumerate}
    \item It refines the class-specific activation mask from a classifier by snapping it to the super-pixels.
    \item It extracts more structural information by training an additional classifier with pure gray images.
    %\item Its training procedure incorporates affine transformation and randomly set input image to grey scale.
    \item The CRF process in the testing phase is regulated by the classifier.
\end{enumerate}

Experimental results show that our proposed system outperforms the baseline by a large margin, and its performance is comparable with the state-of-the-art.

\section{Related work}
\label{sec:related work}

Weakly supervised semantic segmentation has been extensively studied to relieve the data deficiency problem.  According to the types of annotation required by the overall system, existing weakly-supervised methods are based on various problem settings.

For example, user scibble \cite{tang2018regularized,tang2018normalized} and abundant web images \cite{shen2017weakly,shen2018bootstrapping} are considered to regulate the training process or enrich the training data.
Assisted by bounding box, \cite{dai2015boxsup} iterates between automatically generating region proposals and training convolutional networks alternatively.
With pre-trained object detector to reduct the accumulated error, Qi \textit{et al.} \cite{qi2016augmented} proposes a training algorithm that progressively improves segmentation performance with augmented feedback in iterations.  Saleh \textit{et al.} \cite{saleh2018incorporating} extracts the foreground/background mask from a classifier, then obtains multi-class masks by fusing them with information extracted from a weakly-supervised localization network. Isola \textit{et al.} \cite{isola2014crisp} considers crisp boundary detection as higher-order terms for CRF refinement.

A large group of works incorporate supervised pre-trained saliency detector to first generate good foreground/background masks, then use them to obtain class-specific segmentation masks \cite{oh2017exploiting,chaudhry2017discovering,wei2017stc,hou2017bottom,wei2017object,ge2018multi,
huang2018weakly,wei2018revisiting,wang2018weakly,fan2018associating,sun2019saliency}.
As mentioned in the previous section, in the popular semantic segmentation dataset PASCAL VOC 2012 \cite{everingham2010pascal,tang2018regularized}, over 63\% of the images contain one class, where the saliency is equivalent to the semantic segmentation mask, and in all the cases, precise saliency can at least offer precise background mask.  With the help of a well-trained saliency detector, these methods significantly outperform those trained with image labels only.  However, most of the saliency detectors are trained in a fully-supervised manner, and the pixel-wise foreground annotations are still costly to produce.

The challenge of obtaining the semantic segmentation mask with only image label supervision is addressed in different ways. A mutli-step pipeline \cite{shimoda2016distinct,shimoda2018weakly} is proposed to extract class specific maps from a classifier by 1) modifying the activation map calculation, 2) subtracting the activation maps of other classes from that of the target class, 3) aggregating multiple-scale, and 4) augmenting training data for only good cases. Kim \textit{et al.} \cite{kim2017two} tries to get more complete activation masks through two phases training.  In the second phase, the highly discriminative activations from the first training phase are suppressed, forcing the network to discover the next most important parts.  However, it is not clear why suppressing the activation of one part leads to the discovering of another, since there is no clear competition between them.  In other words, as long as activating more parts helps to reduce the loss, the network should do so, and including more parts can be achieved simply by lowering the threshold.  Roy \textit{et al.} \cite{roy2017combining} specifies a deep architecture that fuses the cues from three distinct computation processes via a conditional random field as a recurrent network aiming at generating a smooth and boundary-preserving segmentation.
Kwak \textit{et al.} \cite{kwak2017weakly} proposes Superpixel Pooling Network (SPN), which utilizes superpixel segmentation of input images as a pooling layout to reflect the low-level image structure for learning and inferring semantic segmentation. AffinityNet \cite{Ahn_2018_CVPR} predicts semantic affinity between a pair of adjacent image coordinates.  The propagation of the activation from local discriminative parts to the entire object is then realized by one random walk with the affinities predicted by AffinityNet.  Kolesnikov \textit{et al.} \cite{kolesnikov2016seed} uses cues from a well-trained classifier as a partial segmentation mask for training, and defines the loss function that contains three terms, i.e., the seeding loss to impose the activation in the cue region, the constrain-to-boundary loss to encourage the consistency of the segmentation masks before and after applying CRF, and the expansion loss, which penalizes the classification error.

Our system adopts the main procedure of \cite{kolesnikov2016seed}, and modifies it by three ways: 1) snapping the cues to the super-pixels in the image, 2) fusing the cues of a color image trained classifier and that of a gray image trained classifier to cover more parts, and 3) using prediction of the classifier to suppress non-exist classes during testing.  Unlike \cite{kwak2017weakly, Ahn_2018_CVPR}, our super-pixels are calculated based on the properties of natural images, and no training is involved in predicting the affinities between pixels.

\section{Proposed method}
\label{sec:proposed method}

\subsection{Problem formulation}
\label{subsec:problem formulation}
Let us model each pixel label as a random variable \(X_i\) associated with pixel \(i\) in image \(\mathbf{I}\).   \(X_i\) can take any one label from a pre-defined set \(\mathcal{L}=\{l_1,l_2, \cdots ,l_K\}\).
The segmentation goal is to obtain labels for all the random variables \(\mathbf{X}=\{X_1,X_2,\cdots,X_N\} \) given observation \(\mathbf{I} \), where \(N\) is the total number of pixels in the image.
Our segmentation network \(\mathcal{F}_{seg}(.) \) is a fully convolutional neural network (FCN) whose outputs \(\mathcal{F}_{seg}(\mathbf{I})= \{\mathbf{f}_1, \mathbf{f}_2, \cdots, \mathbf{f}_K \} \) are \(K\) channel maps~\footnote{resize to that of \(\mathbf{I}\) if needed}.  For a well-trained segmentation network, the value at location \(\mathbf{p}_i\) in the \(j^{th}\) output map represents the probability that pixel \(i\) belongs to class \(j\), i.e., \(\mathbf{f}_j(\mathbf{p}_i) = p(X_i=l_{j}|\mathbf{I}) \).  CRF~\cite{lafferty2001conditional,krahenbuhl2011efficient} is applied to \(\mathcal{F}_{seg}(\mathbf{I})\) for further refinement.  Its unary potential is given by
\begin{equation}
	\psi_u(X_i = l_j) = - \log \mathbf{f}_j(\mathbf{p}_i),
\end{equation}
and pairwise potential is the commonly used
\begin{equation}
\begin{aligned}
\psi_p(i, j) =  & \omega_1 \exp (- \frac{||\mathbf{p}_i-\mathbf{p}_j||^2}{2\sigma_{\alpha}^2} - \frac{||\mathbf{I}_i-\mathbf{I}_j||^2}{2\sigma_{\beta}^2}) \\
                 + & \omega_2 \exp (- \frac{||\mathbf{p}_i-\mathbf{p}_j||^2}{2\sigma_{\gamma}^2}),
\end{aligned}
\end{equation}
where \(\mathbf{p}_i \) and \(\mathbf{I}_i\) are the location and color of pixel \(i\).

\subsection{Proposed system overview}
\begin{figure*}[h]
	\centering
	\includegraphics[width=\textwidth]{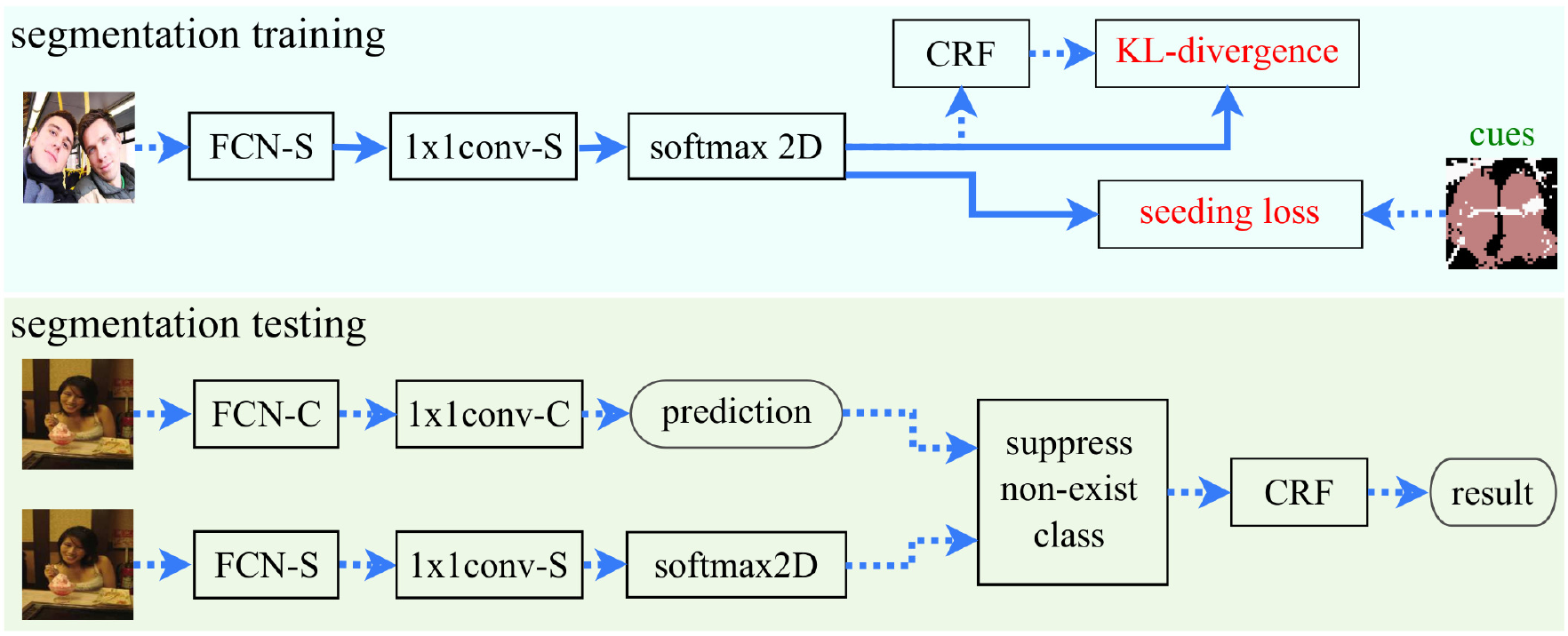}
	\caption{Training and testing the segmentation network. The dashed arrows indicate the processes without backward propagation.}
	\label{fig:seg train test}
\end{figure*}

There are four major steps in our proposed system: 1) classifiers training, 2) cue generation, 3) segmentation network training, and 4) testing.  The first two steps are illustrated in Figure.~\ref{fig:cue generation}, and the last two steps are illustrated in Figure.~\ref{fig:seg train test}.
We first train two classifiers using color images and gray images respectively with multi-label cross entropy loss, then extract a set of cues from each classifier and fuse them.  A set of cues are binary class-specific activation maps that have the same shape as \(\mathcal{F}_{seg}(\mathbf{I})\).  They indicate the confident elements in the coarse segmentation results,  and serve as an incomplete pseudo annotation to train the segmentation network.  Following SEC~\cite{kolesnikov2016seed}, suppose \(C\) is a set of cues, then the seeding loss \(L_{s}\) has the following form:
\begin{equation}
	L_{s}=-\frac{1}{|C|}\sum_{\{i|C(i)=1\}}\log \mathcal{F}_{seg}(\mathbf{I})(i),
\end{equation}
where \(|C|\) is the total number of elements with value 1 in \(C\).

The other loss term we use is the constrain-to-boundary loss \cite{kolesnikov2016seed}.  Let \(Q(\mathbf{I},\mathcal{F}_{seg}(\mathbf{I}))\) denote the result of applying CRF refinement on \(\mathcal{F}_{seg}(\mathbf{I})\), the constrain-to-boundary loss $L_{c}$ is the KL-divergence between \(\mathcal{F}_{seg}(\mathbf{I})\) and \(Q(\mathbf{I},\mathcal{F}_{seg}(\mathbf{I}))\):
\begin{equation}
	 L_{c}=
	 \frac{1}{n} \sum_{i} Q(\mathbf{I},\mathcal{F}_{seg}(\mathbf{I}))(i)\log \frac{Q(\mathbf{I},\mathcal{F}_{seg}(\mathbf{I}))(i)}{\mathcal{F}_{seg}(\mathbf{I})(i)},
\end{equation}
where \(n\) is the total number of elements in \(\mathcal{F}_{seg}(\mathbf{I})\).

As shown in Figure~\ref{fig:seg train test}, in our system, the full loss used to train \(\mathcal{F}_{seg}(.) \) is: \(L = L_s + L_c\).  In the testing phase, the prediction from the color image trained classifier is used to amend the result of the segmentation network \(\mathcal{F}_{seg}(\mathbf{I})\) before CRF refinement.  The key parts of our proposed system are detailed in the following two subsections.

\subsection{Cue Generation}
\label{subsec:cue generation}
\begin{figure*}[h!]
	\centering
  \begin{subfigure}{0.12\textwidth}
		\centering
		\includegraphics[width=\textwidth,height=\textwidth]{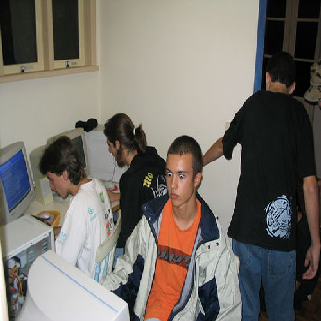}\\
		\includegraphics[width=\textwidth,height=\textwidth]{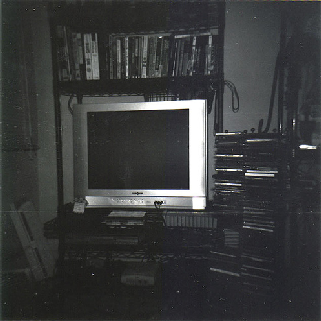}\\
		\includegraphics[width=\textwidth,height=\textwidth]{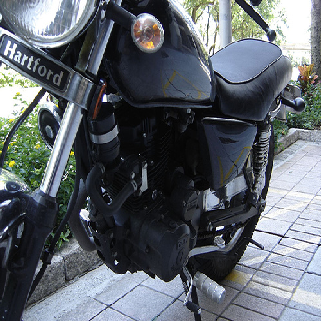}\\
		\includegraphics[width=\textwidth,height=\textwidth]{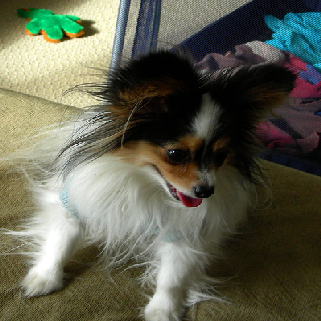}\\
		\includegraphics[width=\textwidth,height=\textwidth]{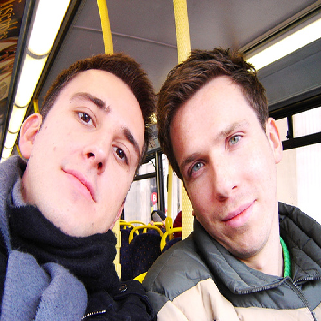}\\
		\includegraphics[width=\textwidth,height=\textwidth]{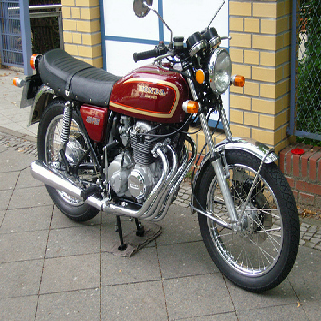}\\
		\includegraphics[width=\textwidth,height=\textwidth]{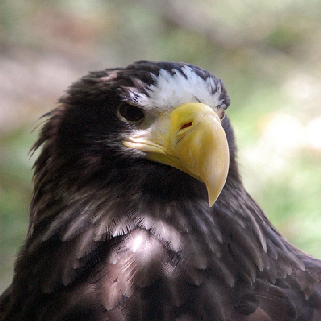}\\
		\includegraphics[width=\textwidth,height=\textwidth]{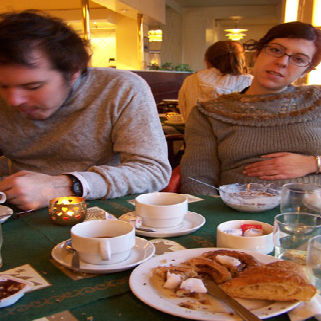}\\
		\includegraphics[width=\textwidth,height=\textwidth]{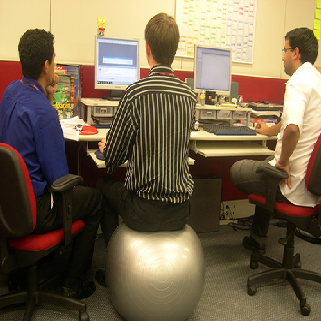}\\
  	\caption{Original}
  	\label{fig:org}
  \end{subfigure}
  \begin{subfigure}{0.12\textwidth}
		\centering
		\includegraphics[width=\textwidth,height=\textwidth]{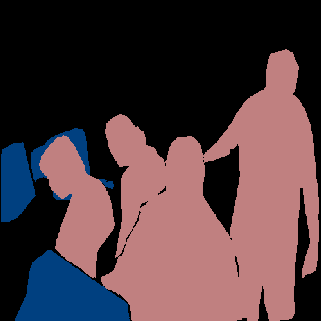}\\
		\includegraphics[width=\textwidth,height=\textwidth]{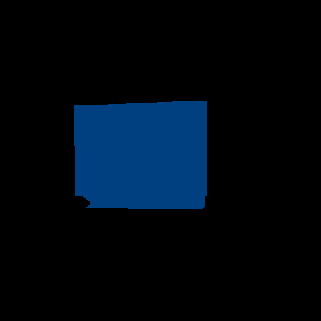}\\
		\includegraphics[width=\textwidth,height=\textwidth]{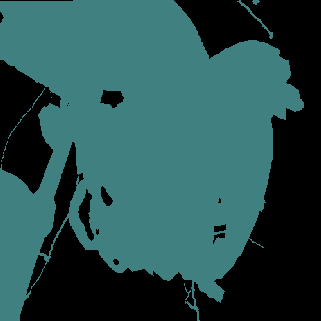}\\
		\includegraphics[width=\textwidth,height=\textwidth]{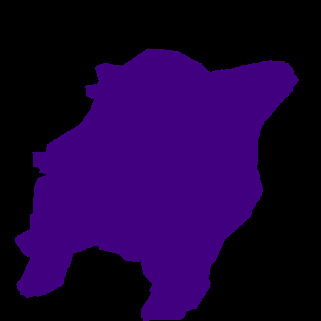}\\
		\includegraphics[width=\textwidth,height=\textwidth]{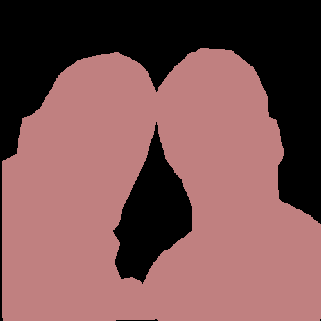}\\
		\includegraphics[width=\textwidth,height=\textwidth]{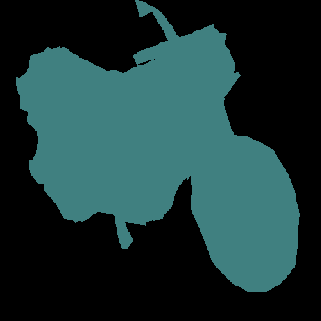}\\
		\includegraphics[width=\textwidth,height=\textwidth]{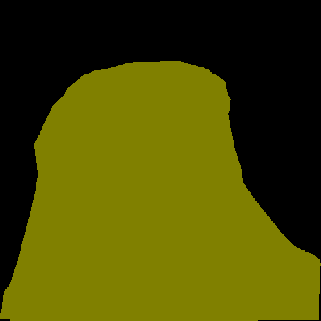}\\
		\includegraphics[width=\textwidth,height=\textwidth]{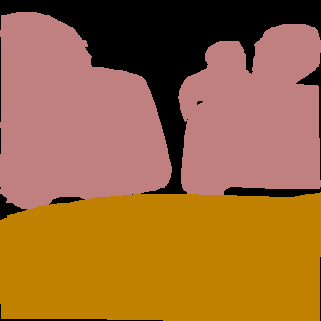}\\
		\includegraphics[width=\textwidth,height=\textwidth]{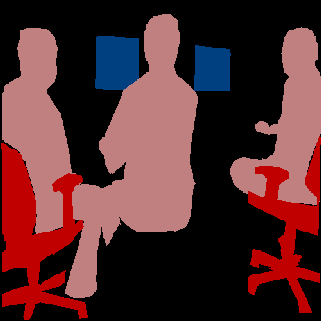}\\
  	\caption{Truth}
  	\label{fig:gt}
  \end{subfigure}
  \begin{subfigure}{0.12\textwidth}
		\centering
		\includegraphics[width=\textwidth,height=\textwidth]{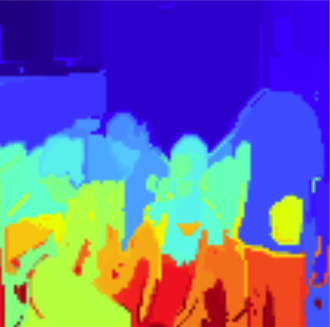}\\
		\includegraphics[width=\textwidth,height=\textwidth]{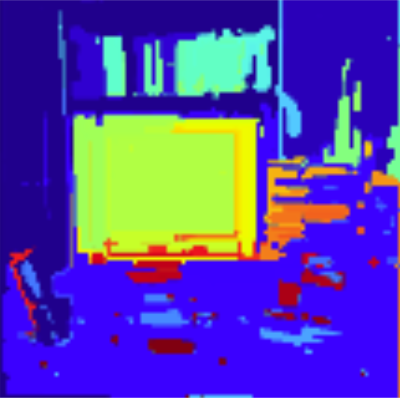}\\
		\includegraphics[width=\textwidth,height=\textwidth]{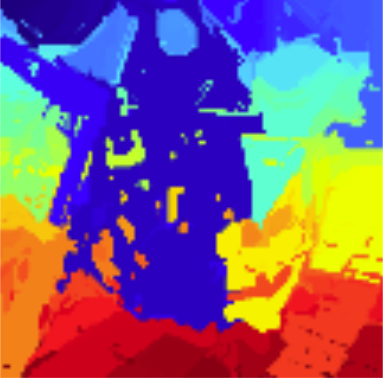}\\
		\includegraphics[width=\textwidth,height=\textwidth]{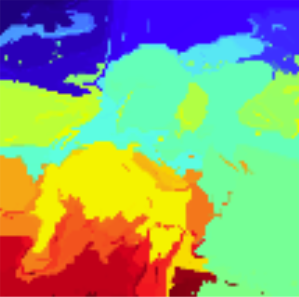}\\
		\includegraphics[width=\textwidth,height=\textwidth]{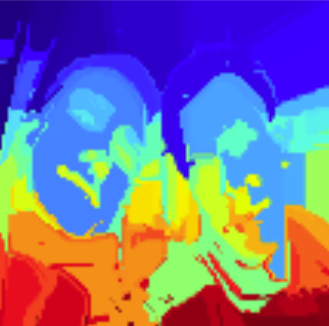}\\
		\includegraphics[width=\textwidth,height=\textwidth]{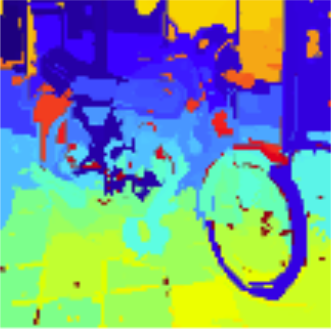}\\
		\includegraphics[width=\textwidth,height=\textwidth]{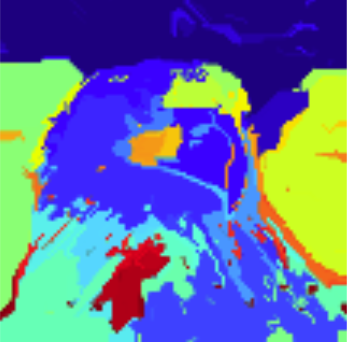}\\
		\includegraphics[width=\textwidth,height=\textwidth]{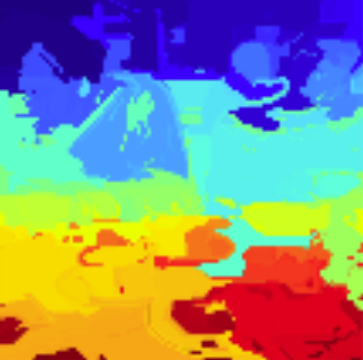}\\
		\includegraphics[width=\textwidth,height=\textwidth]{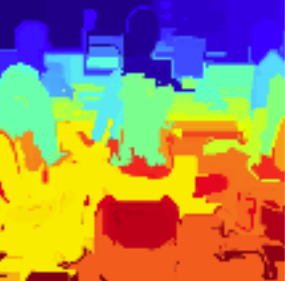}\\
  	\caption{Super pixel}
  	\label{fig:sp}
  \end{subfigure}
	\begin{subfigure}{0.12\textwidth}
		\centering
		\includegraphics[width=\textwidth,height=\textwidth]{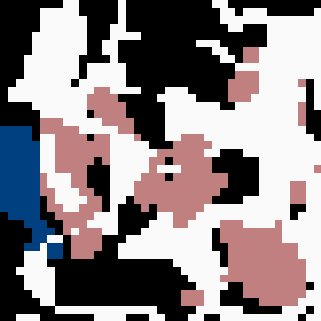}\\
		\includegraphics[width=\textwidth,height=\textwidth]{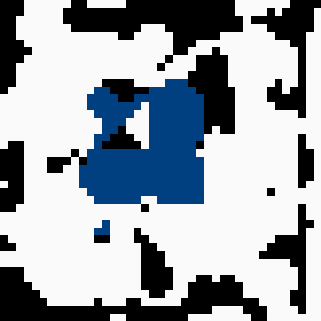}\\
		\includegraphics[width=\textwidth,height=\textwidth]{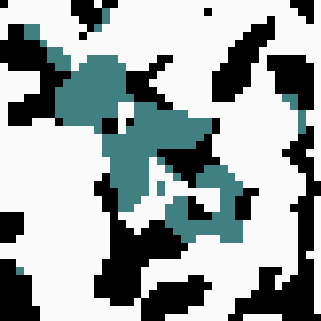}\\
		\includegraphics[width=\textwidth,height=\textwidth]{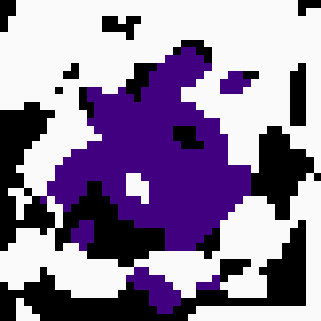}\\
		\includegraphics[width=\textwidth,height=\textwidth]{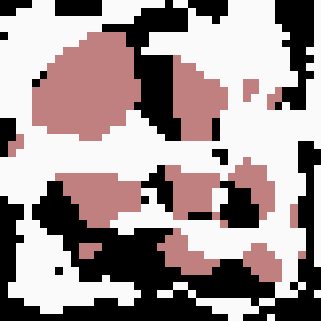}\\
		\includegraphics[width=\textwidth,height=\textwidth]{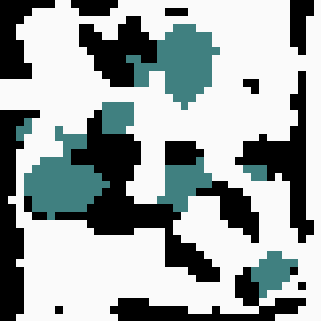}\\
		\includegraphics[width=\textwidth,height=\textwidth]{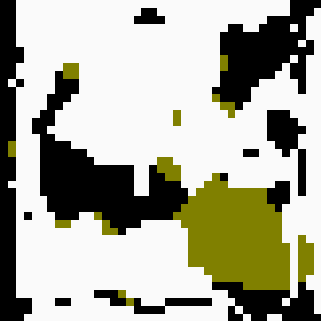}\\
		\includegraphics[width=\textwidth,height=\textwidth]{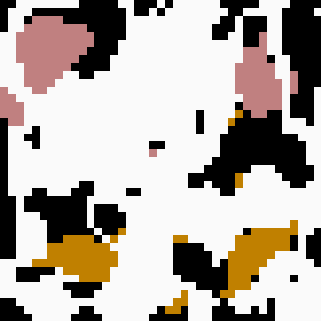}\\
		\includegraphics[width=\textwidth,height=\textwidth]{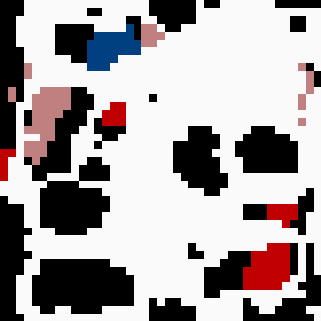}\\
  	\caption{Color cue}
  	\label{fig:cc}
  \end{subfigure}
  \begin{subfigure}{0.12\textwidth}
		\centering
		\includegraphics[width=\textwidth,height=\textwidth]{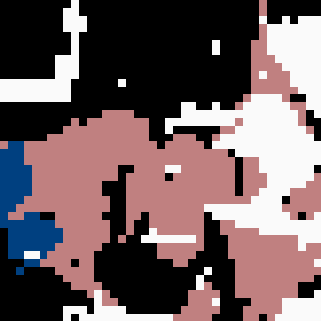}\\
		\includegraphics[width=\textwidth,height=\textwidth]{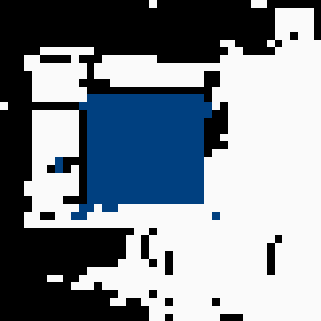}\\
		\includegraphics[width=\textwidth,height=\textwidth]{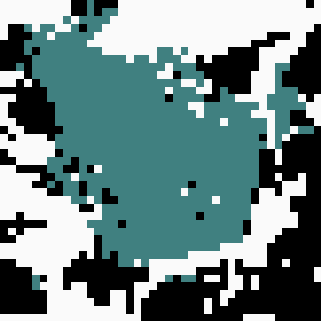}\\
		\includegraphics[width=\textwidth,height=\textwidth]{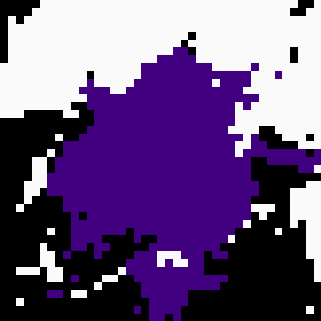}\\
		\includegraphics[width=\textwidth,height=\textwidth]{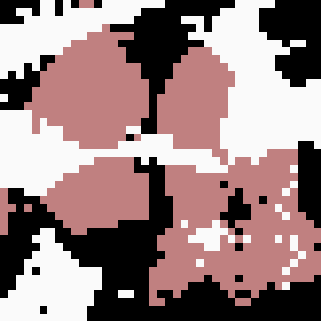}\\
		\includegraphics[width=\textwidth,height=\textwidth]{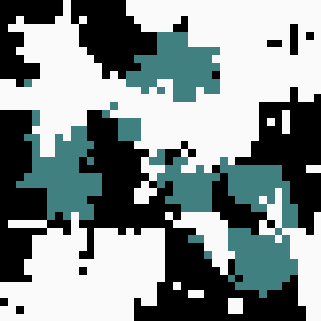}\\
		\includegraphics[width=\textwidth,height=\textwidth]{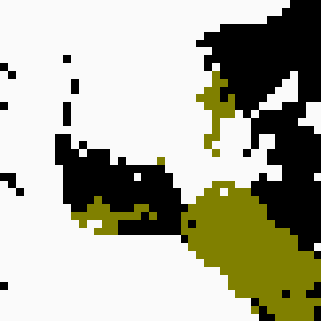}\\
		\includegraphics[width=\textwidth,height=\textwidth]{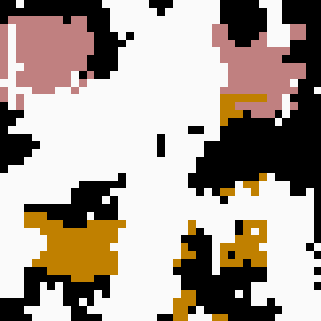}\\
		\includegraphics[width=\textwidth,height=\textwidth]{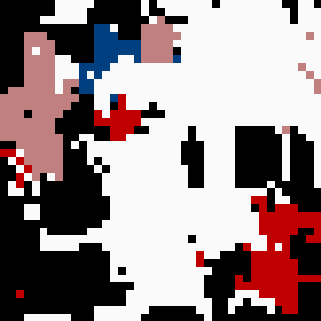}\\
  	\caption{Snapped cc}
  	\label{fig:cc_sp}
  \end{subfigure}
  \begin{subfigure}{0.12\textwidth}
		\centering
		\includegraphics[width=\textwidth,height=\textwidth]{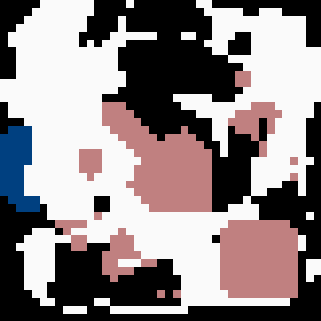}\\
		\includegraphics[width=\textwidth,height=\textwidth]{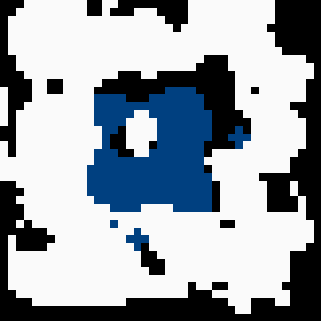}\\
		\includegraphics[width=\textwidth,height=\textwidth]{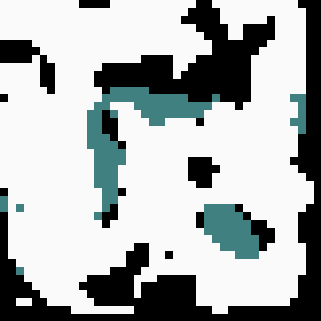}\\
		\includegraphics[width=\textwidth,height=\textwidth]{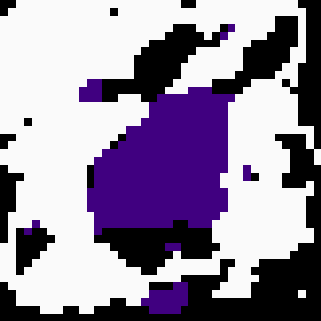}\\
		\includegraphics[width=\textwidth,height=\textwidth]{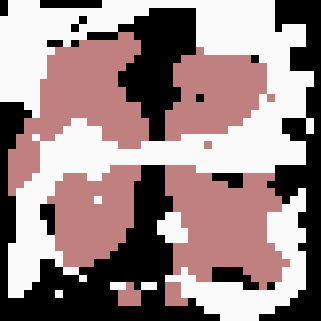}\\
		\includegraphics[width=\textwidth,height=\textwidth]{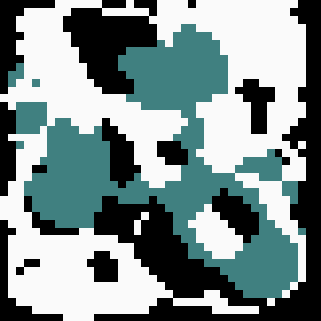}\\
		\includegraphics[width=\textwidth,height=\textwidth]{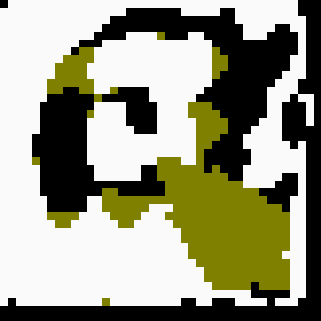}\\
		\includegraphics[width=\textwidth,height=\textwidth]{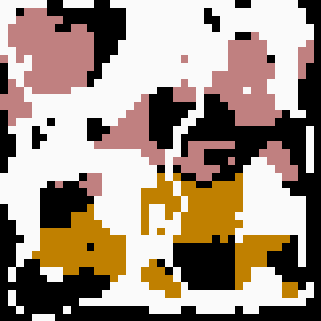}\\
		\includegraphics[width=\textwidth,height=\textwidth]{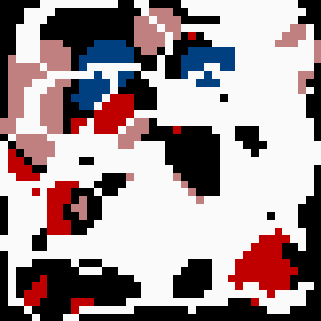}\\
  	\caption{Gray cue}
  	\label{fig:gc}
  \end{subfigure}
	\begin{subfigure}{0.12\textwidth}
		\centering
		\includegraphics[width=\textwidth,height=\textwidth]{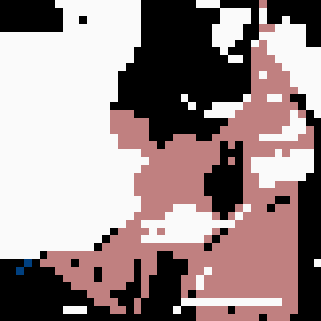}\\
		\includegraphics[width=\textwidth,height=\textwidth]{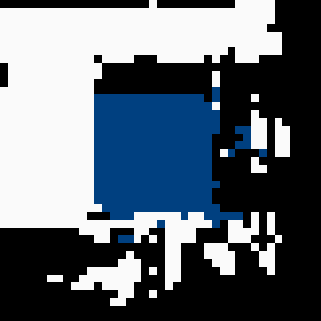}\\
		\includegraphics[width=\textwidth,height=\textwidth]{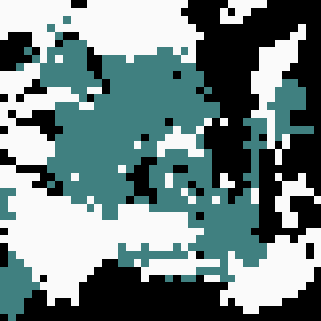}\\
		\includegraphics[width=\textwidth,height=\textwidth]{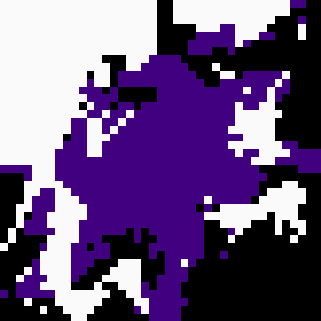}\\
		\includegraphics[width=\textwidth,height=\textwidth]{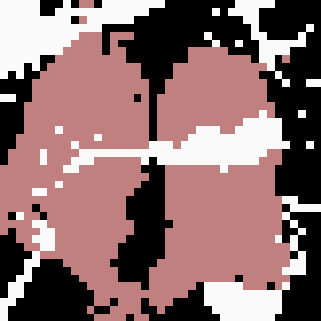}\\
		\includegraphics[width=\textwidth,height=\textwidth]{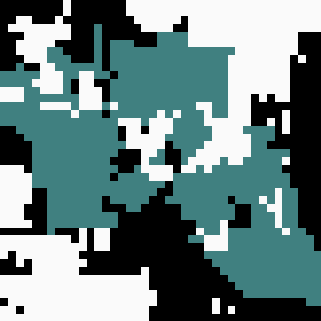}\\
		\includegraphics[width=\textwidth,height=\textwidth]{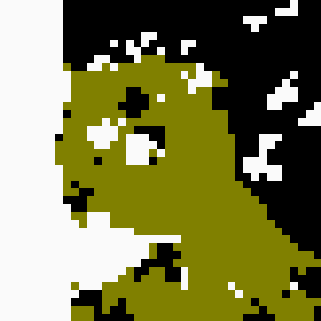}\\
		\includegraphics[width=\textwidth,height=\textwidth]{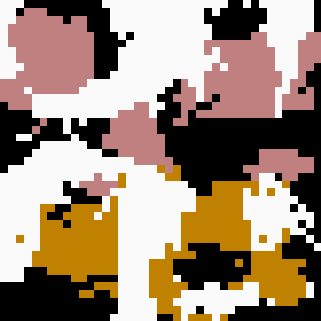}\\
		\includegraphics[width=\textwidth,height=\textwidth]{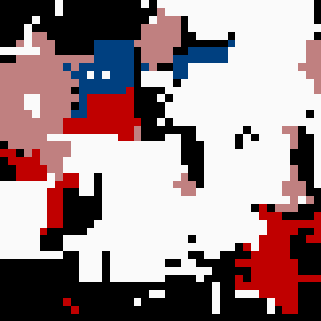}\\
  	\caption{Snapped gc}
  	\label{fig:gc_sp}
  \end{subfigure}
  \begin{subfigure}{0.12\textwidth}
		\centering
		\includegraphics[width=\textwidth,height=\textwidth]{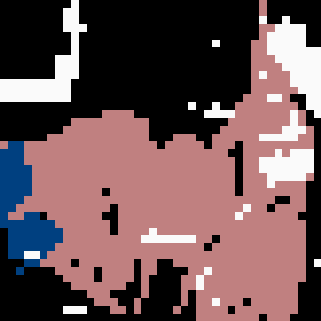}\\
		\includegraphics[width=\textwidth,height=\textwidth]{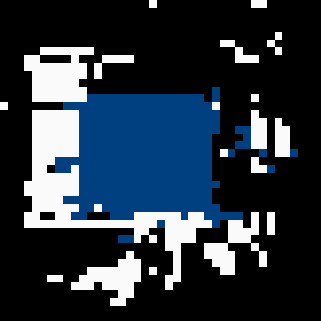}\\
		\includegraphics[width=\textwidth,height=\textwidth]{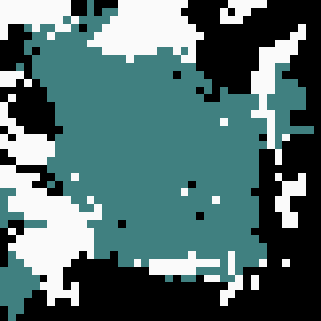}\\
		\includegraphics[width=\textwidth,height=\textwidth]{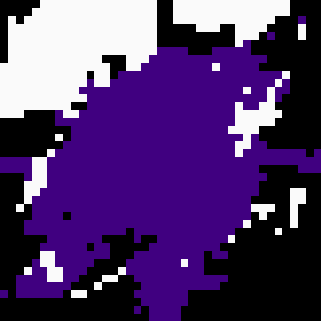}\\
		\includegraphics[width=\textwidth,height=\textwidth]{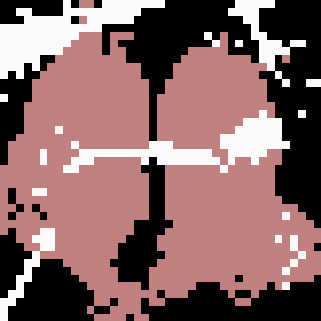}\\
		\includegraphics[width=\textwidth,height=\textwidth]{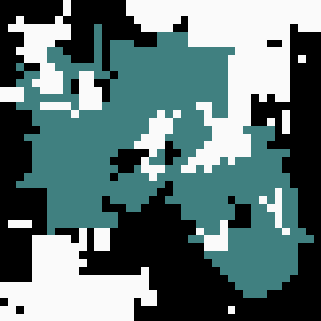}\\
		\includegraphics[width=\textwidth,height=\textwidth]{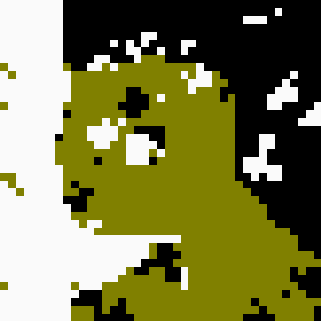}\\
		\includegraphics[width=\textwidth,height=\textwidth]{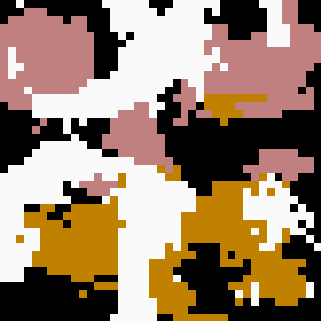}\\
		\includegraphics[width=\textwidth,height=\textwidth]{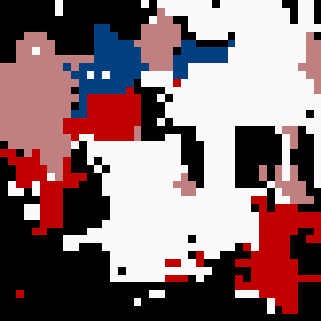}\\
  	\caption{Merged cue}
  	\label{fig:mc}
  \end{subfigure}

	\caption{Illustration of proposed cue generation method.  Each row contains one example and from left to right shows:  \ref{fig:org} original image, \ref{fig:gt} ground truth segmentation mask, \ref{fig:sp} the calculated super-pixels, \ref{fig:cc} the cues extracted from a color image trained classifier (CC), \ref{fig:cc_sp} snapping CC to the super-pixels, \ref{fig:gc} the cues extracted from a gray image trained classifier (GC), \ref{fig:gc_sp} snapping GC to the super-pixels, and \ref{fig:mc} merging cues from snapped CC and snapped GC.
	}
	\label{fig:cue gen}
\end{figure*}

In our system, foreground class-specific activation maps are obtained through CAM~\cite{zhou2016learning}.  We adopt the modification in \cite{shimoda2016distinct,shimoda2018weakly}, i.e. subtracting the activation maps of other classes from that of the target class to obtain clearer separated objects. The binary cues are the pixels above 30\% of the maximum value in an activation map.  To generate the background map, we first summarize all the feature maps of the top two convolutional layers from the classifier, then normalize the result to $[0,1]$.  The pixels with their value smaller than 0.2 are used as background cues.  In Figure~\ref{fig:cue gen} (better viewed in color), some sample images are shown in the first column, each with its ground truth segmentation shown in the second column.  The fourth column contains the binary cues directly obtained from a classifier.  It can be seen that they are coarse, and a large portion of each image is uncovered.
We propose to refine the cues by snapping them to the super-pixels calculated through \cite{felzenszwalb2004efficient}.
An important characteristic of this method is its ability to preserve detail in low-variability image regions while ignoring detail in high-variability regions.  The super-pixels calculated by \cite{felzenszwalb2004efficient} are shown in the third column in Figure~\ref{fig:cue gen}.  We average the cue value in each super-pixel, and binarize the result with a threshold that equals to 0.3 times the maximum value in the mask.  The snapped cues are shown in the fifth column.  Compared with the raw cues, the snapped cues are improved in two ways: 1) the boundaries are better aligned with those of the objects, and 2) both foreground and background cues are spread to cover more complete regions.

\begin{figure*}[h]
	\centering
	\includegraphics[width=\textwidth]{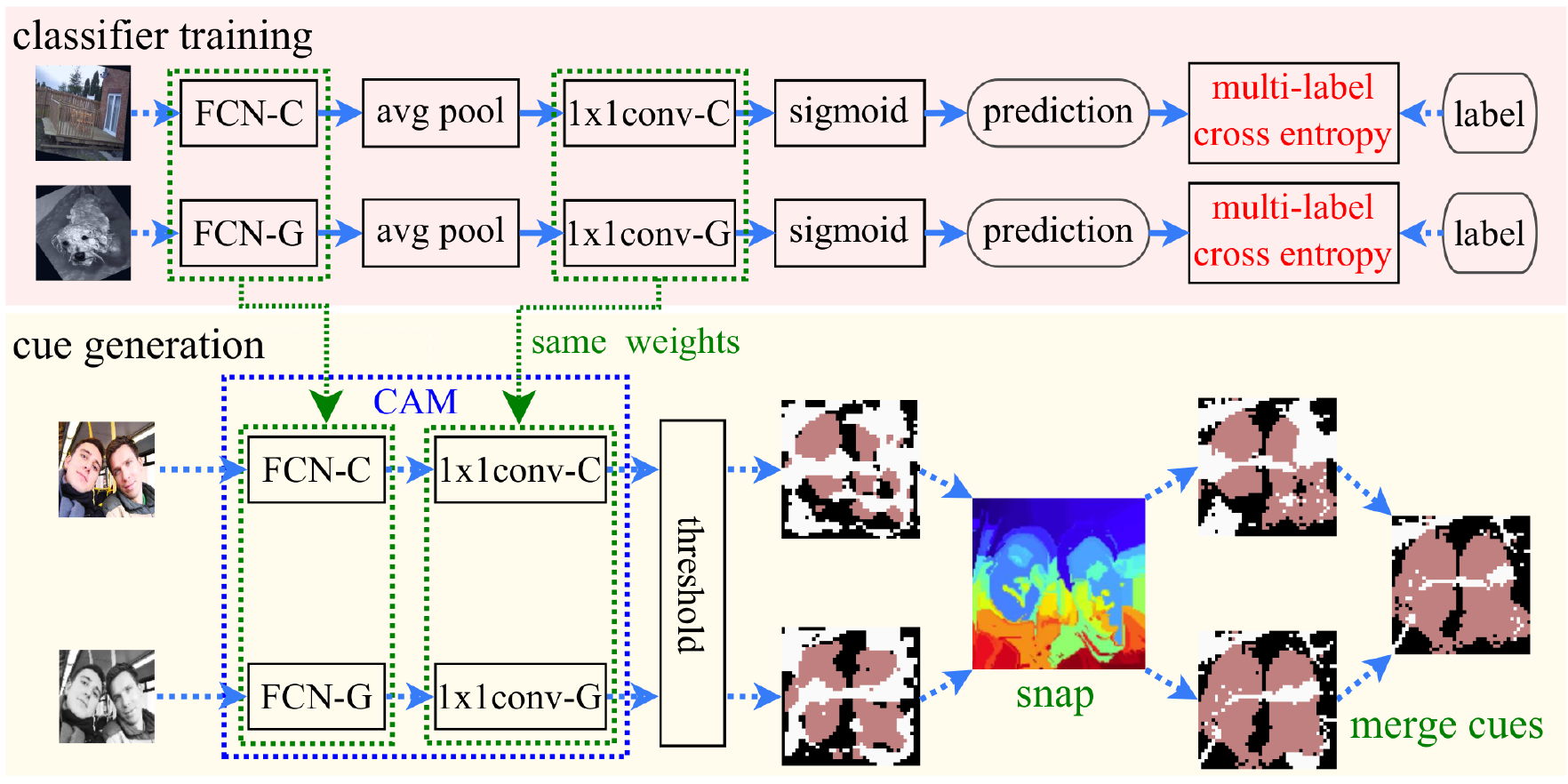}
	\caption{Training classifier and cue generation. The dashed arrows indicate the processes without backward propagation.
	}
	\label{fig:cue generation}
\end{figure*}

As shown in Figure~\ref{fig:cue generation}, another countermeasure we take against the incompleteness of the generated cues is to fuse two sets of cues extracted from a color image trained classifier and a gray image trained classifier.
This procedure is motivated by our two observations: 1) we found that in the PASCAL VOC 2012 dataset \cite{everingham2010pascal,tang2018regularized} some images suffer severe color distortion due to overexposure and extreme light condition, and 2) the class-specific activation maps from a gray image trained classifier can cover more parts that are discriminative in structure but not in color. For example, a color image trained classifier usually ignores the legs of a person since his/her pants can have any color and this high variance information seems not to help the classification task. However, in gray images all people's legs have a similar structure then the network can find these regions are helpful to distinguish people from other classes.  The way we merge two sets of binary cues is simply by applying logical \textit{OR}.  The last column in Figure~\ref{fig:cue gen} shows the merged cues, which are better than both cues from a color image trained classifier~(\ref{fig:cc_sp}) and that from a gray image trained classifier~(\ref{fig:gc_sp})

\subsection{Decouple classification from segmentation}
As shown in the top of Figure~\ref{fig:seg train test}, after the cues are generated, they are used to train the segmentation network.
The loss function in SEC~\cite{kolesnikov2016seed} contains three terms, i.e. the seeding loss imposes the activation in the cue region; the constrain-to-boundary loss, which is the KL-divergence between the outputs of the segmentation network and that of the CRF~\cite{lafferty2001conditional,krahenbuhl2011efficient}; and the expansion loss, which is the multi-label classification error.  In our system, we reserve the seeding loss and constrain-to-boundary loss, but omit the classification error.  The reason for this decouple is that we observe the conflict between the needs of classification and segmentation.   For example, some patterns are shared by multiple classes, e.g., the body fur of cats and dogs, and the wheels of cars and buses, while a classification loss that drives the network to highlight the region that can uniquely identify a class will make the activation maps over discriminative and incomplete for segmentation purpose.

\subsection{Prediction amend testing}
\label{subsec:prediction amend}
We use our previously color image trained classifier in three ways:  1) to generate cues, 2) to initialize the segmentation network, and 3) to amend the result in testing, as shown in the bottom half of Figure~\ref{fig:seg train test}.  Since the cues used to train the segmentation network are incomplete, i.e., only some pixel locations are included, the segmentation network can produce non-exist classes.  The damage of a few wrongly activated pixels can be amplified by CRF, which propagates the unary by location closeness and color similarity.  This problem can be relieved by using the prediction of a classifier to suppress the non-exist classes.  During the testing phase in our system, at each pixel location, the maximum scores among the predicted classes are used as the ceiling value \(c_i = \max_{j | j\in \mathcal{S}_{pred}} \mathbf{f}_j(\mathbf{p}_i)\), and the scores in predicted non-exist classes are suppressed to below this ceiling by a small margin $1e-4$.  The effectiveness of this method is shown in the next section.
\footnote{Notice that \textit{FCN-C, FCN-G} and \textit{FCN-S} are actually the same fully-convolutional structures. \textit{FCN-C} and \textit{FCN-G} are trained from color and gray images respectively. \textit{FCN-S} is trained for the final segementation and initialized from \textit{FCN-C}. It is the same for the \textit{1x1conv} layer which is a 2D convolutional layer with kernel size of $1\times1$.}

\section{Experiments}
\label{sec:experiments}
In this section we validate our proposed system experimentally.

\subsection{Dataset}
Our proposed method is evaluated on the PASCAL VOC 2012 image segmentation benchmark \cite{everingham2010pascal,tang2018regularized}, which has 20 foreground classes and 1 background class.  The dataset contains 1464 images for training, 1449 images for validation and 1456 images for testing.  Following the common practice, we use the augmented training set, i.e., trainaug set, which has 10,582 images, from \cite{hariharan2011semantic}.  Both evaluation and testing results are reported and intermediate results are shown on the evaluation set only since the ground truth segmentation masks for the testing set are not publicly available.  We use the standard PASCAL VOC 2012 segmentation metric mean intersection-over-union (mIoU).

\subsection{System evaluation}
\begin{figure*}[!h]
	\centering
	% \vspace{-2.2cm}
	\includegraphics[width=0.9\textwidth]{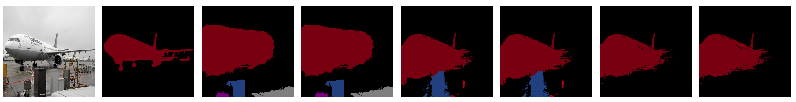}\\
	\includegraphics[width=0.9\textwidth]{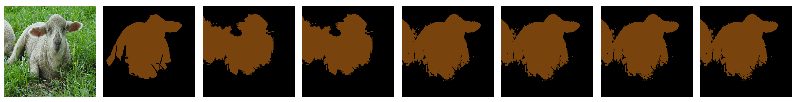}\\
	\includegraphics[width=0.9\textwidth]{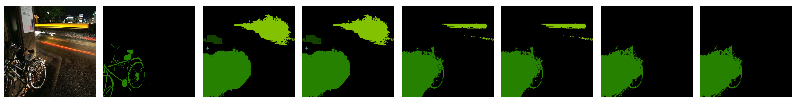}\\
	\includegraphics[width=0.9\textwidth]{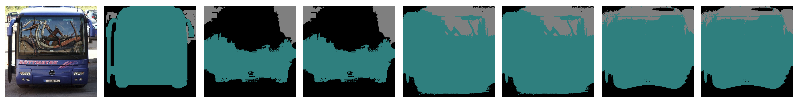}\\
	\includegraphics[width=0.9\textwidth]{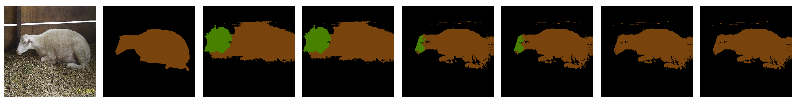}\\
	\includegraphics[width=0.9\textwidth]{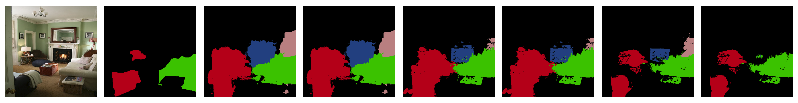}\\
	\includegraphics[width=0.9\textwidth]{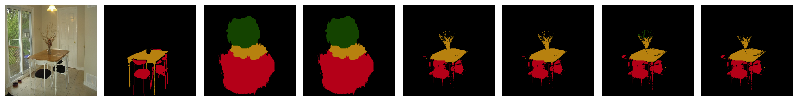}\\
	\includegraphics[width=0.9\textwidth]{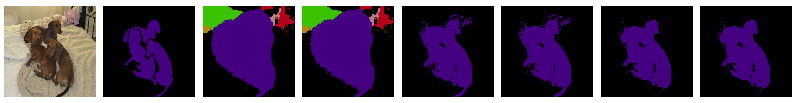}\\
	\caption{
		Qualitative segmentation results of the proposed method.
		From left to right: 1)the original image, 2)the ground truth segmentation, 3)the segmentation results obtained from the top convolutional layer of a classifier, 4)the results of using raw cues to train a segmentation model, 5)using the cues snapped to super-pixels, 6)using the merged cue from both a color image trained classifier and a gray image trained one, 7)using prediction to suppress the non-exist classes (\textbf{proposed}), and 8) using ground truth class label to suppress the non-exist classes (our upper bound).}
		\label{fig:qualitative}
	%\vspace{-0.2cm}
\end{figure*}

The effectiveness of our proposed system can be seen from Table~\ref{tab: sys eval}.  We test our method on two networks, and the detailed network structures can be found in Res38~\cite{wu2019wider,Ahn_2018_CVPR} and Res50~\cite{He_2016_CVPR}.  As described in section~\ref{subsec:problem formulation}, the segmentation results are obtained from the CNN masks, i.e. \(\mathcal{F}_{seg}(\mathbf{I}) \),
and are refined by CRF~\cite{lafferty2001conditional,krahenbuhl2011efficient}.
\textit{CAM} means the \(\mathcal{F}_{seg}(\mathbf{I}) \) is obtained from a classifier through CAM~\cite{zhou2016learning}.
\textit{Raw cue}, \textit{snapped cue} and \textit{merged cue} means the  \(\mathcal{F}_{seg}(\mathbf{I}) \) are the outputs of a segmentation network trained by raw cues, super-pixel snapped raw cues and merged cues respectively.  \textit{With prediction} means the non-existing classes in \(\mathcal{F}_{seg}(\mathbf{I}) \) are suppressed before CRF according to the prediction of a classifier, as described in sub-section~\ref{subsec:prediction amend},
and \textit{with class labels} means using ground truth class labels, which shows the upper-bound performance gain of this process.  \textit{Acc} and
\textit{Rec} means the accuracy and recall of these networks when they are trained as classifiers.  It can be seen that \textit{merged cue} outperforms \textit{snapped cue}, which outperforms both \textit{raw cue} and \textit{CAM}.  This means our proposed cue generation method effectively refines the raw cues and makes them more complete.  Using the predictions from a classifier to amend \(\mathcal{F}_{seg}(\mathbf{I}) \) steadily improve the mIoU.  Qualitative results using Res38~\cite{wu2019wider,Ahn_2018_CVPR} are shown in Figure~\ref{fig:qualitative}.  It can be seen that adopting our proposed cue generation process improve the segmentation results, and using predictions can successfully suppress non-exist classes.

\begin{table*}
\begin{center}
\begin{tabular}{c|| c | c | c | c | c | c || c | c}
     mIoU \% & CAM & raw cue & snapped cue & merged cue & with prediction & with class labels & Acc \% & Rec \% \\
  \hline
 % Vgg16~\footnote{In \cite{kolesnikov2016seed}, the segmentation network and the network used to generate cues are very different, we tried but failed to reproduce the cues with similar quality.  Here we use the cues generated by Res38 to train the segmentation network that has the same structure as in \cite{kolesnikov2016seed}.  These experiments are to shown the improvement of the cues before and after super-pixel snapping and merging, not to compare } & 78 & 90 & 22.8 & 48.2 & 49.7 & 50.0 & 49.5  & 52.6 \\
  Res38 & 40.1 & 49.9 & 55.0 & 55.3 & 56.0 & 57.8 & 91 & 95 \\
  Res50 & 29.5 & 43.6 & 47.9 & 49.3 & 52.4 & 52.4 & 94 & 90
\end{tabular}
\end{center}
\caption{mIoU results on the validation dataset of PASCAL VOC 2012 \cite{everingham2010pascal, mark2015ijcv} segmentation task.  We test our method on two segmentation networks, the names in the first column represent the original networks where they are modified from.  ResNet38 is introduced in \cite{wu2019wider}, and adopted in \cite{Ahn_2018_CVPR}.  Res50 is modified from ResNet-50 in \cite{He_2016_CVPR} by changing the last fully connected layer to a \(1\times1\) convolutional layer. }
\label{tab: sys eval}
\end{table*}

\subsection{Cue evaluation}
One of our main contributions is to modify the raw cues extracted from the classifiers by 1) snapping the cues to super-pixels, and 2) merging the cues from a color image trained classifier and a gray image trained classifier.  In this sub-section, we directly evaluate the cues of training data by mIoU.   Since the cues are partial annotations, the unknown pixels in the cues are treated as background.  Both the mIoU of all the classes and that of the foreground classes are shown in Table~\ref{tab: cue eval}.  It can be seen that our proposed method indeed improve the quality of the cues.

\begin{table}
	\begin{center}
		\begin{tabular}{c| c c c | c c c}
			mIoU \% & raw & snapped & merged & raw & snapped & merged \\
			 \hline
			class   & \multicolumn{3}{c}{all classes} & \multicolumn{3}{c}{foreground classes} \\
			 \hline
			Res38  & 36.5 & 42.9 & 44.9 & 34.7 & 41.3 & 43.3 \\
			Res50  & 33.0 & 42.7 & 49.3	& 31.1 & 41.2 & 48.0
		\end{tabular}
	\end{center}
	\caption{This table shows mIoU evaluation of the cues of the 10,582 trainaug images from \cite{hariharan2011semantic}.  The unknown pixels in the cues are treated as background.  Both the mIoU of all the classes and that of the foreground classes are presented.}
	\label{tab: cue eval}
\end{table}

\subsection{Comparison with other methods}
In this sub-section the proposed method is compared with the latest works under the problem setting that only image labels are used in training, without other annotations through the whole system.  The results on PASCAL VOC 2012 image segmentation benchmark \cite{everingham2010pascal,tang2018regularized} validation data and test data are shown in Table~\ref{tab: compare eval} and Table~\ref{tab: compare test} respectively.  It can be seen that the performance of our proposed method is comparable with that of the state-of-the-art.  The only method outperforming us is \cite{Ahn_2018_CVPR}, which involve training an extra AffinityNet and random walk process.

\begin{table}
	\begin{center}
		\begin{tabular}{c|c c c c c | c c}
		IoU \%& \cite{kolesnikov2016seed} & \cite{roy2017combining} & \cite{kim2017two} & \cite{Ahn_2018_CVPR} & \cite{shimoda2018weakly} & Res38 & Res50 \\
			  \hline
		  bg  & 82.5 & 85.8 & 82.8 & 88.2 & 81.6 & 83.2 & 83.2\\
	aeroplane & 62.9 & 65.2 & 62.2 & 68.2 & 64.9 & 54.8 & 52.0\\
		bike  & 26.4 & 29.4 & 23.1 & 30.6 & 25.8 & 24.5 & 28.8\\
		bird  & 61.6 & 63.8 & 65.8 & 81.1 & 71.4 & 57.7 & 45.4\\
		boat  & 27.6 & 31.2 & 21.1 & 49.6 & 29.2 & 43.5 & 19.2\\
	   bottle & 38.1 & 37.2 & 43.1 & 61.0 & 57.8 & 58.9 & 49.2\\
		bus   & 66.6 & 69.6 & 71.1 & 77.8 & 75.2 & 75.1 & 68.0\\
		car   & 62.7 & 64.3 & 66.2 & 66.1 & 68.0 & 70.2 & 69.0\\
		cat   & 75.2 & 76.2 & 76.1 & 75.1 & 72.7 & 76.8 & 79.0\\
		chair & 22.1 & 21.4 & 21.3 & 29.0 & 15.2 & 26.6 & 24.9\\
		cow   & 53.5 & 56.3 & 59.6 & 66.0 & 46.6 & 68.1 & 58.3\\
 dinningtable & 28.3 & 29.8 & 35.1 & 40.2 & 33.8 & 33.1 & 28.6\\
        dog   & 65.8 & 68.2 & 70.2 & 80.4 & 56.7 & 75.8 & 74.6\\
        horse & 57.8 & 60.6 & 58.8 & 62.0 & 57.1 & 66.9 & 59.7\\
    motorbike & 62.3 & 66.2 & 62.3 & 70.4 & 60.9 & 67.0 & 65.4\\
     person   & 52.5 & 55.8 & 66.1 & 73.7 & 60.7 & 56.1 & 64.1\\
       plant  & 32.5 & 30.8 & 35.8 & 42.5 & 24.1 & 39.1 & 38.5\\
        sheep & 62.6 & 66.1 & 69.9 & 70.7 & 65.4 & 67.1 & 68.1\\
         sofa & 32.1 & 34.9 & 33.4 & 42.6 & 31.5 & 32.5 & 33.2\\
        train & 45.4 & 48.8 & 45.9 & 68.1 & 43.9 & 53.0 & 48.1\\
   TV/monitor & 45.3 & 47.1 & 45.6 & 51.6 & 35.3 & 46.0 & 42.2\\
              \hline
      average & 50.7 & 52.8	& 53.1 & 61.7 & 51.3 & 56.0	& 52.4
		\end{tabular}
	\end{center}
	\caption{IoU results on the validation dataset of PASCAL VOC 2012 \cite{everingham2010pascal, mark2015ijcv} segmentation task.}
	\label{tab: compare eval}
\end{table}

\begin{table}
	\begin{center}
		\begin{tabular}{c|c c c c c | c c}
  IoU \% & \cite{kolesnikov2016seed} & \cite{roy2017combining} & \cite{kim2017two} & \cite{Ahn_2018_CVPR} & \cite{shimoda2018weakly} & Res38 & Res50\\
			  \hline
			  bg  & 83.5 & 85.7 & 83.4 & 89.1 & 83.0 & 83.9 & 83.8\\
	    aeroplane & 56.4  & 58.8 & 62.2 & 70.6 & 67.5 & 56.8 & 52.5\\
			bike  & 28.5 & 30.5 & 26.4 & 31.6 & 29.7 & 24.9 & 29.6\\
			bird  & 64.1 & 67.6 & 71.8 & 77.2 & 69.7& 61.5 & 46.7\\
			boat  & 23.6 & 24.7 & 18.2 & 42.2 & 28.8 & 39.1 & 18.0\\
		   bottle & 46.5 & 44.7 & 49.5 & 68.9 & 59.7 & 56.0 & 55.5\\
			bus   & 70.6 & 74.8 & 66.5 & 79.1 & 71.2 & 75.9 & 70.9\\
			car   & 58.5 & 61.8 & 63.8 & 66.5 & 66.4 & 69.0 & 66.0\\
		    cat   & 71.3 & 73.7 & 73.4 & 74.9 & 69.8 & 71.6 & 77.0\\
			chair & 23.2 & 22.9 & 19.0 & 29.6 & 18.6 & 25.6 & 21.9\\
			cow   & 54.0 & 57.4 & 56.6 & 68.7 & 49.8 & 65.0 & 54.9\\
	 dinningtable & 28.0 & 27.5 & 35.7 & 56.1 & 44.7 & 42.4 & 33.7\\
			dog   & 68.1 & 71.3 & 69.3 & 82.1 & 49.4 & 75.1 & 71.7\\
			horse & 62.1 & 64.8 & 61.3 & 64.8 & 60.5 & 67.4 & 60.0\\
	    motorbike & 70.0 & 72.4 & 71.7 & 78.6 & 73.5 & 74.7 & 72.3\\
		 person   & 55.0 & 57.3 & 69.2 & 73.5 & 61.8 & 54.7 & 66.5\\
		   plant  & 38.4 & 37.0 & 39.1 & 50.8 & 32.7 & 45.6 & 43.7\\
			sheep & 58.0 & 60.4 & 66.3 & 70.7 & 62.7 & 66.2 & 62.3\\
			 sofa & 39.9 & 42.8 & 44.8 & 47.7 & 39.0 & 39.2 & 36.1\\
			train & 38.4 & 42.2 & 35.9 & 63.9 & 34.3 & 52.1 & 43.1\\
	   TV/monitor & 48.3 & 50.6 & 45.5 & 51.1 & 36.5 & 44.8 & 41.7\\
			  \hline
		  average & 51.7 & 53.7	& 53.8 & 63.7 & 52.8 & 56.7 & 52.8
		\end{tabular}
	\end{center}
	\caption{IoU results on the test dataset of PASCAL VOC 2012 \cite{everingham2010pascal, mark2015ijcv} segmentation task.}
	\label{tab: compare test}
	\vspace{-5mm}
\end{table}

\section{Conclusion}
\label{sec:conclusion}
In this paper, we propose a novel weakly-supervised semantic segmentation method using image-level labels only.  The cues extracted from the well-trained classifiers are used as incomplete annotations to train a segmentation network.  We propose to modify the raw cues by 1) snapping them to the super-pixels; 2) merging two sets of cues from a color image trained classifier and the cues from a gray image trained classifier.  Both qualitative results and numeric results show that the proposed two processes effectively improve the quality of the cues.  We also propose to decouple the classification loss from the loss to train a segmentation network since they have a conflict over the shared pattern regions.  During the testing, we propose to use the prediction from the previously trained classifier to suppress the non-existing classes in the segmentation network output before CRF refinement, and this method steadily improves the mIoU.  The performance of the overall proposed system is comparable with that of the state-of-the-art.

%\newpage

\end{document}